\pdfoutput=1

\documentclass[11pt]{article}

\usepackage[final]{acl}
\usepackage{mathtools}
\usepackage{tikz}
\usetikzlibrary{positioning, arrows.meta, shapes.geometric}

\usepackage[inline]{enumitem}

\usepackage{xcolor}

\newcommand*\circled[1]{\texttt{#1}}
\newcommand*\circledt[1]{\texttt{#1}}

\newcommand*\lbs[1]{\underline{#1}}
\newcommand*\obs[1]{\underline{\textbf{#1}}}

\usepackage{xspace}

\newcommand{\lvbabbr}[0]{\emph{lvb}}
\newcommand{\itaabbr}[0]{\emph{ita}}

\newcommand{\RNmodels}[0]{{\texttt{R}- and \texttt{N}-models}}
\newcommand{\Nmodels}[0]{{\texttt{N}-models}}
\newcommand{\Rmodels}[0]{{\texttt{R}-models}}
\newcommand{\Lmodels}[0]{{\texttt{L}-models}}
\newcommand{\latoit}[0]{{\lvbabbr{} $\rightarrow$ \itaabbr{}}}
\newcommand{\ittola}[0]{{\itaabbr{} $\rightarrow$ \lvbabbr{}}}

\newcommand{\testsetone}[0]{{Testset~1}} 
\newcommand{\testsettwo}[0]{{Testset~2}} 
\newcommand{\testsetthree}[0]{{Testset~3}} 

\usepackage{enumitem}
\usepackage{rotating}
\usepackage{tabularx}
\usepackage{listings}
\lstdefinestyle{json}{
    basicstyle=\ttfamily\small,
    showstringspaces=false,
    breaklines=true,
    frame=lines,
    breakindent=0em,
    columns=flexible,
    keepspaces=true,
    literate=
     *{0}{{{\color{blue}0}}}{1}
      {1}{{{\color{blue}1}}}{1}
      {2}{{{\color{blue}2}}}{1}
      {3}{{{\color{blue}3}}}{1}
      {4}{{{\color{blue}4}}}{1}
      {5}{{{\color{blue}5}}}{1}
      {6}{{{\color{blue}6}}}{1}
      {7}{{{\color{blue}7}}}{1}
      {8}{{{\color{blue}8}}}{1}
      {9}{{{\color{blue}9}}}{1}
      {:}{{{\color{purple}:}}}{1}
      {,}{{{\color{purple},}}}{1}
      {\{}{{{\color{darkgray}\{}}}{1}
      {\}}{{{\color{darkgray}\}}}}{1}
      {[}{{{\color{darkgray}[}}}{1}
      {]}{{{\color{darkgray}]}}}{1}
      {’}{{{\color{black}'}}}{1}
      {á}{{\'a}}{1}
      {é}{{\'e}}{1}
      {ó}{{\'o}}{1}
      {í}{{\'i}}{1}
      {è}{{\`e}}{1}
      {ë}{{\"e}}{1}
      {ü}{{\"u}}{1}
      {ö}{{\"o}}{1}, 
}

\usepackage{times}
\usepackage{latexsym}

\usepackage[T1]{fontenc}

\usepackage[utf8]{inputenc}

\usepackage{microtype}

\usepackage{inconsolata}

\usepackage{graphicx}

\title{Rule-Based, Neural and LLM Back-Translation:\\Comparative Insights from a Variant of Ladin}

\author{Samuel Frontull \and Georg Moser \\
     Department of Computer Science \\ University of Innsbruck, Innsbruck, Austria \\ {\tt \{samuel.frontull, georg.moser\}@uibk.ac.at}}

\begin{document}
\maketitle
\begin{abstract}
This paper explores the impact of different back-translation approaches on machine translation for Ladin, specifically the Val Badia variant. Given the limited amount of parallel data available for this language (only 18k Ladin--Italian sentence pairs), we investigate the performance of a multilingual neural machine translation model fine-tuned for Ladin--Italian. In addition to the available authentic data, we synthesise further translations by using three different models: a fine-tuned neural model, a rule-based system developed specifically for this language pair, and a large language model. Our experiments show that all approaches achieve comparable translation quality in this low-resource scenario, yet round-trip translations highlight differences in model performance.
\end{abstract}

\section{Introduction}

In recent years, a variety of methods have been developed to apply neural machine translation (NMT) also in low-resource scenarios~\cite{Shi:etal:2022,Hadd:etal:2022,Rana:etal:2023}. 
The back-translation technique has shown to be particularly effective in such settings~\cite{Senn:etal:2016,Edun:etal:2018}, 
offering the potential for substantial improvements in translation quality.

This work investigates the influence of the back-translation model selection for a low-resource language. We do this, by comparing the results obtained by fine-tuning a pre-trained multilingual NMT model using synthesised translations generated by 
\begin{enumerate*}[label=(\roman*)]
  \item a NMT system fine-tuned on the available parallel data,
  \item a rule-based machine translation (RBMT) system developed for the specific language pair, and
  \item a large language model (LLM) prompted to translate given texts, accompanied by 8 exemplary samples.
\end{enumerate*}

The quality of the synthesised data, which in turn is determined by the underlying models used to generate it, matters~\cite{Burl:Yvon:2018}.
In our case, the synthesised translations originate from three models based on a different paradigm.
Thus, the synthesised data is characterised by the specific strengths and weaknesses of the respective paradigms. 

Rule-based systems are robust and computationally lightweight, but may face challenges in dealing with ambiguity. Moreover, they lag behind at the grammatical level. 
Neural models show a high ability to adapt to provided texts, but perform less well when confronted with out-of-domain data~\cite{Shen:etal:2021}. 
In contrast, language model-based approaches (LLMs) are praised for their ability to produce fluent, coherent texts, but they are prone to hallucinations~\cite{Rawte:etal:2023}. 
It is therefore an interesting question to investigate how this affects the quality of the NMT models trained on this data. 
This comparative analysis sheds light on the nuanced contrasts inherent in the different MT methods.

Our results show that in this low-resource scenario the back-translation model does not have a significant impact, and the performance of the models converges to similar results in terms of BLEU/chrF++ points.
This assertion is supported by an empirical analysis carried out on the Val Badia variant of Ladin.
Our main contributions are:

\begin{itemize}
  \item we are the first to explore MT for Ladin in general, with a specific focus on the Val Badia variant,
  \item we compare an RBMT-, an NMT- and an LLM-based back-translation, providing insights into the efficacy of the methods for Ladin,
  \item we establish baseline results and make the test data, the RBMT system, as well as the best-performing models publicly available
\end{itemize}

In Section~\ref{sec:data} we describe our data collection and corpus creation process.
In Section~\ref{sec:models} we present the three different methods used for the back-translation of monolingual Ladin data into Italian.
Section~\ref{sec:experiments} gives an overview of the conducted experiments and Section~\ref{sec:results} presents the obtained results.
Section~\ref{sec:related-work} discusses related work and similar approaches. In Section~\ref{sec:conclusion}, we summarize and discuss future work.

\paragraph{The Ladin Language}

Ladin is an officially recognised minority language, and thus taught in schools, used in the media, and employed in public administration. 
For this reason, an effective machine translation system could make a significant contribution to facilitating and supporting communication in this language.
However, Ladin is still an unexplored language in the field of machine translation.
Indeed, nearly no parallel data\footnote{In a machine readable format.} is publicly available for this language, except for a few hundred samples on OPUS~\cite{Tied:2012}.
This language, spoken by around 30,000 people in the northern Italian Dolomite regions, exhibits significant diversity across its five main variants (\emph{Val Badia}, \emph{Gherdëina}, \emph{Fascia}, \emph{Fodom}, \emph{Anpezo}), each shaped uniquely by its development in different valleys. This diversity is not only evident in the spoken language but has also resulted in distinct standards for written communication.
The first author of the paper originates from the \emph{Val Badia} and is a native speaker of Ladin. Therefore, in this work we concentrate on the standard written language of this valley.
In the rest of the paper, we will use \lvbabbr{} as language code to refer to this variant of Ladin, and \itaabbr{} for Italian.

\section{Data}
\label{sec:data}

This section gives an overview of the linguistic resources available for the Ladin language and describes the method employed to collect data for the specific Val Badia variant of Ladin.

\subsection{Available Resources}
Publicly accessible parallel data for Ladin is scarce. 
The Open Parallel Corpus~\cite{Tied:2012} e.g. lists 1543 Ladin--German, 220 Ladin--Italian, and 81 Ladin--English sentences.
However, these texts are mainly specific to the variants of Gherdëina and Fassa and were not disseminated by public institutions.
For our experiments, we were provided with the archive of the weekly newspaper \emph{La Usc di Ladins}\footnote{\url{https://www.lausc.it/}} and a digitised version of the dictionary Ladin Val Badia -- Italian~\cite{Moli:etal:2016}. 
From these data sources we extracted monolingual texts as well as a small dataset of parallel sentences. 
We furthermore used the dictionary as the basis for implementing a RBMT system.
The collection of other parallel texts is time-consuming and has therefore been left for future work.

\subsection{Parallel Data}

The Ladin (Val Badia) -- Italian dictionary~\cite{Moli:etal:2016} contains, alongside the word entries, also sentences that illustrate their usage. 
For these sentences the corresponding Italian translation is also given. 
We have collected this data to create our training dataset, which contains a total of $18,139$ sentences. 
These sentences are basic and short because they were created specifically to illustrate the use of words and phrases. 
The average length is $23.43$ and $25.69$ characters for Ladin and Italian respectively. 
This dataset has been publicly released.\footnote{\url{https://www.doi.org/10.57967/hf/1878}}

\subsection{Ladin Monolingual Data}

The Ladin newspaper \emph{La Usc di Ladins}, digitally archived since 2012, provides an extensive dataset of monolingual texts.
These texts are published in five different variants, each corresponding to one of the five Ladin valleys.
We extracted these texts from the PDF documents and segmented them into individual sentences using the NLTK library~\cite{Bird:etal:2009}, specifically setting Italian as the language to accommodate Ladin.
In total, we accumulated $1,937,608$ sentences. 
These sentences had to be categorised by variant, as described below.

\paragraph{Variant Classification}
In order to train a variant classifier, labeled training data is essential. However, the monolingual data from the newspaper PDFs lacked these labels. Therefore, we collected the texts from the newspaper's website.\footnote{\url{https://www.lausc.it}} Here, the article excerpts are categorized according to their origin valleys and the corresponding language variants, allowing us to create a labeled dataset.

We gathered a corpus of $7,766$ article excerpts with a total of $42,745$ individual sentences for training. These sentences were then split into training (comprising $75\%$ of the sentences) and test data (the remaining $25\%$). Using the $2,500$ most frequent $3$-gram characters as features, we trained an XGBoost variant-classifier~\cite{Chen:Gues:2016}. On the test data, our classifier achieved $94.48\%$ accuracy in classifying these 5 labels.

The resulting model was used to predict the variant of each of the $1,937,608$ sentences in the monolingual dataset.
Table~\ref{tab:articles-distribution} reports the respective number of classifications (and the total number of characters) for each variant.
746,704 sentences were classified as \texttt{val-badia} and were considered for further processing.

\begin{table}[t]
    \centering
    \begin{tabular}{lcc}
    \hline
    \textbf{variant} & \textbf{\# sentences} & \textbf{\# characters}\\
    \hline
    val-badia & $746.704$ & $71.619.515$ \\
    gherdeina & $491.575$ & $57.704.414$ \\
    fascia & $407.605$ & $52.504.357$\\
    fodom & $146.049$ & $16.615.059$ \\
    anpezo & $145.674$ & $16.425.301$ \\
    \hline
    \end{tabular}
    \caption{ \label{tab:articles-distribution}
    Variant classification of monolingual data.
    }
\end{table}

\paragraph{Data Preparation}
Because of the spelling reform in 2015, we further processed the sentences classified as \texttt{val-badia} to exclude any with words that are no longer valid.
To do this, we used the implementation of our RBMT system which is explained in more detail in Section~\ref{subsec:rbmt}.
We used the system to identify unknown words and tried to adapt them to the new spelling according to certain rules. 
Sentences where this was not possible were left out. 
This process ensured that the filtered sentences fully adhered to the new spelling, which also facilitated the rule-based translations.
We collected a total of 274,665 sentences ($\approx 31\%$ of the extracted sentences) which constitute the monolingual Ladin data we used in our experiments.
Among the unused sentences, $\approx 100k$ contain only one unknown word/typo so there would be still  potential to acquire additional data if additional time were spent analysing and preparing these texts.

\subsection{Italian Monolingual Data}
As monolingual data for Italian, we used the \texttt{ELRC-CORDIS\_News} dataset\footnote{\url{https://elrc-share.eu/}} from OPUS~\cite{Tied:2012}, which contains 123,691 Italian sentences.

\subsection{Test Data} \label{sec:testsets}

This section introduces the three test sets on which the models were evaluated. 
This test data differs considerably from the training data, so that it can be considered out-of-domain data.

\paragraph{\testsetone{}}
This dataset includes the statute of the \emph{Stiftung Südtiroler Sparkasse}, a nonprofit foundation dedicated to supporting and promoting various initiatives and projects, primarily within the province of Bolzano. The document is rich in formal and legal terminology. It contains 424 sentences\footnote{\url{https://huggingface.co/datasets/sfrontull/stiftungsparkasse-lld_valbadia-ita}}.

\paragraph{\testsettwo{}}
This dataset is a festive compendium of the history of the region associated with this language~\cite{Kager:2022}. It combines historical narratives with legal and administrative statements. The result is a mixture of stylistic elements and lexical domains. It contains 833 sentences\footnote{\url{https://huggingface.co/datasets/sfrontull/autonomia-lld_valbadia-ita}}.

\paragraph{\testsetthree{}} This dataset delves into the literary realm with the classic story of Pinocchio~\cite{Coll:2017}, a text rich in narrative prose, dialogue and idiomatic expressions, challenging the models with its creative and figurative language. It contains 1563 sentences\footnote{\url{https://huggingface.co/datasets/sfrontull/pinocchio-lld_valbadia-ita}}.

\section{Back-translation Strategies}
\label{sec:models}

The so-called \emph{back-translation}, first introduced in~\citet{Senn:etal:2016}, refers to the process of automatic translation of monolingual texts in the target language to the source language. This method of enriching additional training data in the source-to-target translation direction (where the target side remains authentic) has proven to be particularly effective and is particularly valuable in low-resource scenarios.
In this section we present the three different back-translation strategies used in our research to translate monolingual Ladin texts into Italian.

\subsection{Neural MT}

There is evidence that low-resource languages benefit from multilingual models~\cite{Ahar:etal:2019}.
For this reason, we opted to utilise a pre-trained, multilingual model, specifically the \texttt{Helsinki-NLP/opus-mt-ine-ine}\footnote{\url{https://huggingface.co/Helsinki-NLP/opus-mt-ine-ine}} model available from the Hugging Face Model Hub, as our base model. 
This model, which is part of OPUS-MT~\cite{Tied:Thot:2020}, was trained to translate between 135 Indo-European languages, to which Ladin and Italian also belong. 

The Marian MT model, configured for \texttt{Helsinki-NLP/opus-mt-ine-ine}, features 6 encoder and 6 decoder layers, each with 8 attention heads and a feed-forward dimension of 2048. The model employs a beam search size of 6, a dropout rate of 0.1, and an embedding size of 512. It shares embeddings between the encoder and decoder.

We fine-tuned this model for the two translation directions \latoit{} and \ittola{} on the available authentic training data. We trained a single model for both directions by using the tags \texttt{>>ita<<} for \latoit{} and \texttt{>>lld\_Latn<<}
\footnote{We (re)used the tag \texttt{>>lld\_Latn<<} because it is listed as a valid target language ID, as few Ladin texts were already included in the training of this model.} 
for the opposite direction as prefixes of the source text.
In the rest of the paper, we refer to this fine-tuned model as \circledt{N1}. 
For fine-tuning, we utilized the AdamW optimizer\footnote{\url{https://huggingface.co/docs/transformers/v4.41.0/en/main\_classes/optimizer\_schedules\#transformers.AdamW}} with the defaults settings.

The fine-tuning greatly improves the model in both translation directions, as the scores reported in Table~\ref{tab:results-ita-lvb} and~\ref{tab:results-lvb-ita} show. 
This demonstrates that the data is reliable and that the model adapts well.

\subsection{Rule-based MT} \label{subsec:rbmt}
For low-resource languages, RBMT frameworks offer a crucial advantage: leveraging linguistic expertise to overcome the limitations of data-driven methods~\cite{Khan:etal:2021}.
Considering the similar sentence structure and composition of Ladin and Italian (they are both Romance languages), 
it can be assumed that a rule-based MT system can also perform well without excessive structural transfer work.
The available Ladin Val Badia-Italian dictionary served as the foundation for the rule-based MT system we developed in Apertium~\cite{Forc:Tyer:2016} for this language pair.

This dictionary provides, in addition to the individual words and word translations, also a list of all inflected forms for each lemma.
To effectively utilise this dictionary within our translation system, we mapped the lexicographical data to paradigms within the framework of Apertium (monodix format). 
Specifically, we created 742 paradigms for a total of 19,034 lemmas.
This extensive set includes multiple lexical categories: 597 adverbs, 3,366 adjectives, 11,496 nouns, 162 pronouns, and 2,439 verbs. 
Additionally, we incorporated proper nouns, short phrases, and wordgrams that were identified during the monolingual text extraction process. 
The resulting bilingual dictionary contains a total of 30,468 entries.
The integration with Apertium was facilitated by connecting to and reusing the pre-existing module for Italian\footnote{\url{https://github.com/apertium/apertium-ita}}. 
The Ladin module\footnote{\url{https://github.com/schtailmuel/apertium-lld-ita}} and the Ladin--Italian\footnote{\url{https://github.com/schtailmuel/apertium-lld}} module can be found on GitHub.
In the rest of the paper, we refer to this RBMT system as \circledt{R1}.

According to \texttt{aq-covtest}\footnote{\url{https://wikis.swarthmore.edu/ling073/Apertium-quality}} \circledt{R1} has a coverage of $96.66\%$ on \testsetone{}, $95.81\%$ on \testsettwo{} and $95.90\%$ on \testsetthree{}. 
However, since we did not develop disambiguation modules, we designed the system to select the first suggestion in cases of morphological and lexical ambiguity, which can sometimes result in incorrect choices that may distort the meaning of the texts.
To counteract this, and to further enhance the rule-based translation system, we extracted the $900$ most common word $n$-grams from the texts and added their corresponding translations as entries to the bilingual dictionary.

In addition to the data, we have also included 13 1-level structural transfer rules to avoid common errors. For example, in Ladin, the word \emph{pa} is used to emphasize a question. In Italian, however, there is no corresponding word for this purpose. We have therefore developed a rule to exclude this word from the translation. The other rules include gender correction, dealing with reflexive verbs and prepositions.

\subsection{MT with a Large Language Model}

LLMs have shown remarkable capability in understanding and generating human-like text across various languages and domains~\cite{Brow:etal:2020}.
However, their performance in MT tasks exhibits significant variability across languages, especially when comparing high-resource languages to low-resource languages~\cite{Robi:etal:2023}.
We explore the utilisation of a LLM, specifically \emph{GPT-3.5 Turbo}~\cite{openai:2024}, to generate translations from Ladin to Italian. 
The process involved leveraging the advanced capabilities of the LLM, accessed through the \texttt{gpt-3.5-turbo-0125} API endpoint.
In the rest of the paper, we refer to this LLM as \circledt{L1}.

To enhance throughput and reduce the number of API requests, we generated the translation of 16 Ladin texts in a single request.
We provided a set of 8 example translations in JSON format, randomly selected from the available authentic training data and instructed the LLM to generate translations for 16 Ladin texts, which were also provided as a JSON dictionary, with empty Italian translations. 
Listing~\ref{lst:llm-prompt} (Appendix~\ref{sec:appendix-llm}) showcases an exemplary prompt. 

With this prompting approach, we translated the entire monolingual Ladin corpus into Italian. By providing the exemplary translations as JSON, we were able to reduce the failure rate (invalid/incomplete answers). The extent to which these examples also helped with the translation itself remains open.
The entire process spanned approximately 100 hours, with an average processing time of around 22 seconds per request.

\section{Experiments}
\label{sec:experiments}

We used the \texttt{opus-mt-ine-ine}\footnote{\url{https://huggingface.co/Helsinki-NLP/opus-mt-ine-ine}} model as base model for the experiments.
In the rest of the paper, we use \circledt{BM} to refer to this model.
We fine-tuned \circledt{BM} with the various data sets using the Transformers library~\cite{Wolf:etal:2020}, specifically leveraging the \texttt{Seq2SeqTrainer} module. 
We always trained a single model for both directions using the corresponding prefixes.

We configured the training to process batches of 16 samples, and restricted the input and output sequences to a maximum of 128 tokens to ensure manageable computation loads. 
The models were evaluated each 16,000 steps.
As a stopping criterion, we used three consecutive evaluations resulting in an improvement of less than $0.2$ chrF points on the validation set.
For training, we utilised an NVIDIA TITAN RTX graphics card with 24 GB. 
In total, we have trained 15 models:

\begin{itemize}
\item Model \circledt{N1}: \circledt{BM} fine-tuned with the available parallel data consisting of 18,139 sentences.
\item Models \circledt{N2}/\circledt{R2}/\circledt{L2}: \circledt{BM} fine-tuned with authentic data and Ladin monolingual data backtranslated (BT) to Italian using \circledt{N1}/\circledt{R1}/\circledt{L1} respectively.
\item Models \circledt{N3}/\circledt{R3}/\circledt{L3}: This iteration extends the training base of \circledt{N2}/\circledt{R2}/\circledt{L2} by integrating Italian monolingual data that has been translated into Ladin utilising \circledt{N2}/\circledt{R2}/\circledt{L2} respectively.
\item Models \circledt{N4}/\circledt{R4}/\circledt{L4}: \circledt{BM} fine-tuned with same training data as \circledt{N3}/\circledt{R3}/\circledt{L3} models, but with Ladin and Italian monolingual data backtranslated with \circledt{N3}/\circledt{R3}/\circledt{L3} model.
\item Models \circledt{N5}/\circledt{R5}/\circledt{L5}: This iteration extends the training base of \circledt{N4}/\circledt{R4}/\circledt{L4} by adding also the forward-translations (FT) as training data.
\item Models \circledt{A1}/\circledt{A2}: \circledt{A1} was trained on the combined training data used to trained \circledt{N4}, \circledt{R4} and \circledt{L4}. In \circledt{A2} we additionally included the forward-translations into the training data.
\end{itemize}

\begin{table}[t]
  \centering 
  \begin{tabularx}{0.4\textwidth}{Xrrr}
  \textbf{} & \textbf{\testsetone{}} & \textbf{\testsettwo{}} & \textbf{\testsetthree{}}\\\hline
  \textit{ref} & $425.7$ & $306.3$ & $697.4$ \\ \hline
  \circled{BM} & $545.8$ & $325.8$ & $595.7$ \\ 
  \circled{N1} & $1237.6$ & $437.8$ & $805.3$ \\ 
  \circled{N2} & $633.3$ & $414.0$ & $695.1$ \\
  \circled{N3} & ${484.5}$ & $331.8$ & $606.8$ \\
  \circled{N4} & $367.5$ & $323.4$ & $605.4$ \\ 
  \circled{N5} & $476.2$ & ${320.9}$ & ${593.4}$ \\  \hline

  \circled{R1} & $559.5$ & $421.5$ & $727.8$ \\ 
  \circled{R2} & $593.8$ & $405.7$ & $722.2$ \\ 
  \circled{R3} & $434.8$ & $309.5$ & $601.0$ \\ 
  \circled{R4} & $402.4$ & ${305.3}$ & ${594.3}$ \\ 
  \circled{R5} & ${387.8}$ & $306.1$ & $608.1$ \\ \hline

  \circled{L1} & ${{380.3}}$ & $294.3$ & ${{517.8}}$\\
  \circled{L2} & $695.6$ & ${{406.1}}$ & $675.3$ \\
  \circled{L3} & $396.0$ & $345.4$ & $634.4$ \\
  \circled{L4} & $377.0$ & $318.0$ & $563.7$\\
  \circled{L5} & $393.3$ & $316.1$ & $569.6$ \\\hline
  \end{tabularx}
\caption{\label{tab:perplexity-ita}  
Mean perplexity (\itaabbr{}) of selected models.
}
\end{table}

We refer to the models \circledt{N1}, \dots, \circledt{N5} that were trained with NMT backtranslated data as \Nmodels{}. Analogously, we use the term \Rmodels{} and \Lmodels{} to refer to RBMT and LLM models, respectively.

Models \circledt{N4}/\circledt{R4}/\circledt{L4} illustrate the gains achieved through iterative back-translation~\cite{Hoan:etal:2018}. Additionally, models \circledt{N5}/\circledt{R5}/\circledt{L5} demonstrate potential improvements achievable with synthetically generated forward-translation data.

We evaluated these models on the 3 test sets presented in Section~\ref{sec:testsets}. 
The results are presented and analysed in the following section.

\begin{table*}[t]
    \centering 
    \begin{tabularx}{\textwidth}{Xcrrr}
    & & \textbf{\testsetone{}} & \textbf{\testsettwo{}} & \textbf{\testsetthree{}}\\
    \textbf{Ladin (Val Badia) $\rightarrow$ Italian}  & & \small{BLEU / chrF++} & \small{BLEU / chrF++} & \small{BLEU / chrF++}\\
    \hline
    NMT \footnotesize{\texttt{opus-mt-ine-ine}} & \circled{BM} 
      & $8.17/34.81$ & $8.07/34.27$ & $2.29/21.12$\\
    \circled{BM} fine-tuned with authentic data & \circled{N1} 
      & $12.65/41.55$ & $11.49 / 39.90$ & $11.83/36.40$ \\
    \; $+$ \lvbabbr{} monolingual BT with \circled{N1} &  \circled{N2} 
      & $13.01/42.98$ & $12.40/41.26$ & $13.23/36.84$  \\
    \;\quad $+$ \itaabbr{} monolingual BT with \circled{N2} &  \circled{N3} 
      & $21.98/50.32$ & $19.37/47.35$ & $15.01/39.15$  \\
    \; $+$ \lvbabbr{} and \itaabbr{} monolingual BT with \circled{N3}  &  \circled{N4} 
      & $\lbs{22.90}/\lbs{50.67}$ & $\lbs{21.12}/\lbs{48.38}$ & $\lbs{16.17}/\lbs{40.41}$ \\
      \;\quad $+$ \lvbabbr{} and \itaabbr{} monolingual FT with \circled{N3}  &  \circled{N5} 
      & $21.49/49.94$ & $20.53/48.16$ & $15.10/39.47$ \\\hline
    RBMT \footnotesize{\texttt{apertium-lld-ita}} & \circled{R1} 
      & $11.38/39.72$& $11.60/41.49$ & $8.48/34.48$  \\
    \circled{BM} fine-tuned with authentic data &  &  \\
    \; $+$ \lvbabbr{} monolingual BT with \circled{R1} &  \circled{R2} 
      & $14.43/42.76$ & $13.27/42.00$ & $13.99/37.37$    \\
    \;\quad $+$ \itaabbr{} monolingual BT with \circled{R2} &  \circled{R3} 
      & $22.17/50.33$ & $19.27/48.17$ & $15.89/40.19$ \\
    \; $+$ \lvbabbr{} and \itaabbr{} monolingual BT with \circled{R3} &  \circled{R4} 
      & $21.36/50.24$ & $20.27/\lbs{49.08}$ & $16.34/\obs{40.76}$ \\
    \;\quad $+$ \lvbabbr{} and \itaabbr{} monolingual FT with \circled{R3} &  \circled{R5} 
      & $\lbs{22.50}/\lbs{50.64}$ & $\lbs{20.37}/49.04$ & $\obs{16.36}/40.47$ \\\hline
    LLM \footnotesize{\texttt{gpt-3.5-turbo-0125}} &   \circled{L1} 
      & $\underline{\textbf{26.77}}/\underline{\textbf{53.20}}$ & $21.17/48.52$& $10.37/32.36$ \\
    \circled{BM} fine-tuned with authentic data & & \\
    \; $+$ \lvbabbr{} monolingual BT with \circled{L1} &  \circled{L2} 
      & $12.93/43.20$ & $12.21/41.21$ & $13.22/36.94$  \\
    \;\quad $+$ \itaabbr{} monolingual BT with \circled{L2} &  \circled{L3} 
      & $22.69/50.74$ & $20.37/48.40$ & $\lbs{15.26}/38.99$   \\
    \; $+$ \lvbabbr{} and \itaabbr{} monolingual BT with \circled{L3} &  \circled{L4} 
      & $23.01/51.17$ & $\lbs{21.38}/\lbs{49.24}$ & $15.12/\lbs{39.37}$ \\
    \;\quad $+$ \lvbabbr{} and \itaabbr{} monolingual FT with \circled{L3} &  \circled{L5} 
      & $23.11/50.84$ & $20.86/48.50$ & $15.19/39.29$ \\\hline
    ALL \circled{BM} fine-tuned with authentic data & & \\
    \; $+$ \lvbabbr{} and \itaabbr{} monolingual BT with \circled{N3}, \circled{R3}, \circled{L3} &  \circled{A1} 
      & $23.58/50.68$ & $21.30/48.78$ & $15.32/39.56$ \\
    \;\quad $+$ \lvbabbr{} and \itaabbr{} monolingual FT with \circled{N3}, \circled{R3}, \circled{L3} &  \circled{A2} 
      & $\lbs{24.12}/\lbs{51.42}$ & $\obs{22.24}/\obs{49.69}$ & $\lbs{15.98}/\lbs{39.64}$ \\
    \end{tabularx}
  \caption{\label{tab:results-lvb-ita}  
  Evaluation Results for Ladin to Italian Translation
  }
\end{table*}

\section{Results and Discussion} 
\label{sec:results}

The results of the various experiments conducted are presented in Table~\ref{tab:results-lvb-ita} and Table~\ref{tab:results-ita-lvb}, where the SacreBLEU~\cite{Post:2018} and chrF++~\cite{Popo:2015} scores for different models and test sets are detailed. 
To facilitate comparison, the best scores for each approach have been underlined, and the overall best scores for each testset are highlighted in bold.

Additionally, as recommended in~\citet{Edun:etal:2020}, in Table~\ref{tab:perplexity-ita} we report the mean perplexity values for the Italian translations generated by the different models to complement BLEU's emphasis on adequacy. Perplexity measures how well a language model can predict the next word in a sequence based on the preceding words. Lower perplexity means that the model is more confident and accurate in its predictions, indicating that it can better reproduce the structure and patterns of the language it generates. 
Therefore, we present the mean perplexity values obtained from GPT-2 ~\cite{Radf:etal:2019} , computed using the implementation available from Hugging Face\footnote{\url{https://huggingface.co/spaces/evaluate-metric/perplexity}}.

Several findings can be deduced from these results, and will be discussed below.
In general, there is evidence that augmenting the training data with monolingual data through back-translation is effective.
\circledt{N1} shows that fine-tuning the model with only authentic training data substantially improves the results in both directions (compared to \circledt{BM}) in terms of BLEU/chrF++ points. This shows on the one hand that the training is effective and on the other hand that the available data is adequate. However, it is also evident that the model generates less fluent text, as indicated by the perplexity scores which increase for this model. 

The results reveal a progression in difficulty among the test sets, where \testsetthree{} emerges as the most challenging one. 
On this test set, all approaches achieve similar low scores, suggesting the presented approach may face limitations with more complex texts.

\begin{table*}[t]
    \centering 
    \begin{tabularx}{\textwidth}{Xcrrr}
    & & \textbf{\testsetone{}} & \textbf{\testsettwo{}} & \textbf{\testsetthree{}}\\
    \textbf{Italian $\rightarrow$ Ladin (Val Badia)}  & & \small{BLEU / chrF++} & \small{BLEU / chrF++} & \small{BLEU / chrF++}\\
    \hline
    NMT \footnotesize{\texttt{opus-mt-ine-ine}} & \circled{BM} 
      & $0.08/5.34$ & $0.55/13.68$ & $0.05 / 6.86$ \\
    \circled{BM} fine-tuned with authentic data & \circled{N1} 
      & $10.22/37.11$ & $10.14 / 37.48$& $12.76/35.31$    \\    %
    \; $+$ \lvbabbr{} monolingual BT with \circled{N1} &  \circled{N2} 
     & $19.09/46.92$ & $18.05/45.44$ & $16.50/37.46$    \\
    \;\quad $+$ \itaabbr{} monolingual BT with \circled{N2} &  \circled{N3} 
      & $19.54/\lbs{47.02}$ & $\lbs{19.45}/\lbs{46.21}$ & $\obs{16.66}/37.36$ \\
    \; $+$ \lvbabbr{} and \itaabbr{} monolingual BT with \circled{N3}  &  \circled{N4} 
      & $19.61/46.35$ & $19.16/45.63$ & $16.40/\lbs{37.84}$ \\
      \;\quad $+$ \lvbabbr{} and \itaabbr{} monolingual FT with \circled{N3}  &  \circled{N5} 
      & $\lbs{20.24}/46.72$ & $19.39/45.88$ & $15.56/36.97$ \\\hline

    RBMT \footnotesize{\texttt{apertium-lld-ita}} & \circled{R1} 
      & $4.94/37.50$ & $4.50 / 36.89$& $3.19/27.44$    \\
    \circled{BM} fine-tuned with authentic data &  &  \\
    \; $+$ \lvbabbr{} monolingual BT with \circled{R1} &  \circled{R2} 
      & $19.18/46.59$ & $16.96/44.97$ & $15.21/36.76$    \\
    \;\quad $+$ \itaabbr{} monolingual BT with \circled{R2} &  \circled{R3} 
      & $19.86/46.83$ & $17.70/45.69$ & $15.04/36.60$    \\
    \; $+$ \lvbabbr{} and \itaabbr{} monolingual BT with \circled{R3} &  \circled{R4} 
      & $\lbs{20.93}/\lbs{47.65}$ & $\lbs{19.32}/\lbs{46.58}$ & $\lbs{16.65}/\obs{38.16}$ \\
      \;\quad $+$ \lvbabbr{} and \itaabbr{} monolingual FT with \circled{R3} &  \circled{R5} 
      & $19.97/46.88$ & $18.65/46.19$ & $16.61/38.12$ \\\hline

    LLM \footnotesize{\texttt{gpt-3.5-turbo-0125}} &   \circled{L1} 
      & $5.54/29.03$ & $3.84 / 28.98$& $1.16/18.60$ \\
    \circled{BM} fine-tuned with authentic data & & \\
    \; $+$ \lvbabbr{} monolingual BT with \circled{L1} &  \circled{L2} 
      & $\obs{22.09}/\obs{48.69}$ & $19.71/46.59$ & $14.16/35.67$ \\
    \;\quad $+$ \itaabbr{} monolingual BT with \circled{L2} &  \circled{L3} 
      & $21.59/48.23$ & $\obs{19.96}/\obs{49.96}$ & $14.23/35.81$  \\
    \; $+$ \lvbabbr{} and \itaabbr{} monolingual BT with \circled{L3} &  \circled{L4} 
      & $20.82/47.86$ & $19.87/46.59$ & $\lbs{16.55}/\lbs{38.04}$ \\
      \;\quad $+$ \lvbabbr{} and \itaabbr{} monolingual FT with \circled{L3} &  \circled{L5} 
      & $20.93/47.70$ & $19.38/46.37$ & $15.84/37.29$ \\\hline
    ALL \circled{BM} fine-tuned with authentic data & & \\

    \; $+$ \lvbabbr{} and \itaabbr{} monolingual BT with \circled{N3}, \circled{R3}, \circled{L3} &  \circled{A1} 
      & $19.83/47.16$ & $\lbs{19.94}/\lbs{46.40}$ & $\lbs{16.54}/\lbs{37.91}$ \\
    \;\quad $+$ \lvbabbr{} and \itaabbr{} monolingual FT with \circled{N3}, \circled{R3}, \circled{L3} &  \circled{A2} 
      & $\lbs{20.81}/\lbs{47.50}$ & $19.71/46.36$ & $16.36/37.82$ \\
    \end{tabularx}
  \caption{\label{tab:results-ita-lvb}  
  Evaluation Results for Italian to Ladin Translation
  }
\end{table*}

In the translation direction \latoit{}, the best results were achieved by combining the different back-translations, as model \circledt{A2} results indicate. This emphasises the importance of a broad and diversified dataset. Remarkably, the \circledt{A2} model is also competitive in the reverse translation direction (\itaabbr{} to \lvbabbr{}), although it does not achieve the best results. 

A comparison of the models \circledt{N1}, \circledt{R1} and \circledt{L1} suggests that the LLM generates more fluent texts (low perplexity) but perhaps does not always accurately reproduce the meaning, as attested by the performance on \testsetthree{} (low perplexity but also low BLEU score). In this assessment, the RBMT system \circledt{R1} also performs better than the fine-tuned NMT model \circledt{N1}. One of the reasons for the high perplexity values of \circledt{N1} is that this model tends to hallucinate because it has been fine-tuned with a small data set. However, this does not seem to affect the performance as the models trained on this data do not perform considerably worse.

The performance of the LLM varies significantly, with pronounced differences between the three test sets in both directions of translation. The significant difference observed between \testsetone{} and \testsettwo{} in the translation direction from \latoit{} cannot be seen in the \RNmodels{}. It remains unclear to what extent the LLM benefits from the given examples in the prompt. However, by providing an example, the propensity for errors was minimised, resulting in fewer mistakes during execution.
Even though LLMs are not (yet) suitable for generating texts in low-resource languages out-of-the-box (see performance of \circledt{L1} in Table~\ref{tab:results-ita-lvb}), Ladin and low-resource languages in general could benefit from this technology. Our experiments show that models trained on back-translations from \circledt{L1} performed best on \testsetone{} and \testsettwo{} in the translation direction \ittola{}.

The inclusion of forward translations in the training data did not consistently improve the models, with the exception of the \texttt{R}-models for \latoit{}. This suggests that these synthesised texts introduce too much noise. However, model \circledt{A2} was able to benefit from this data in the translation direction \latoit{}. Filtering this data could slightly improve the model.

As the models achieve similar scores on the test data, we also examined the quality of round-trip translations to gain additional insights. For this, we used 10k sentences from the monolingual Ladin and Italian data (which were also used for training, hence the high scores), translated them into the other language and then back-translated them. This concept of so-called \emph{round-trip translation} is a suitable evaluation method ~\cite{Zhuo:etal:2023}. We used the \circledt{R4}/\circledt{N4}/\circledt{L4} models for this purpose, applying one model $A$ for one direction and the same or a different model $B$ for the opposite direction. Table~\ref{tab:results-rtt} shows the obtained results. It can be clearly seen that the results are worse when a different model is used for the reverse translation. This shows that although the models achieve similar results with the test data, they work differently. The \circledt{R4} model proves to be the most stable here, as its translations can be back-translated well by all three models. For other combinations, a high variance can be observed.

The translation models \circledt{N4}\footnote{\url{https://doi.org/10.57967/HF/2695}}, \circledt{R4}\footnote{\url{https://doi.org/10.57967/HF/2693}}, and \circledt{L4}\footnote{\url{https://doi.org/10.57967/HF/2694}} have been released on Hugging Face, making them accessible for further research and application.

\begin{table}[t]
    \centering 
    \begin{tabularx}{.45\textwidth}{ccc}
    $A/B$ & \textbf{\lvbabbr{} $\xrightarrow[]{A}$ \itaabbr{} $\xrightarrow[]{B}$ \lvbabbr{}} & \textbf{\itaabbr{} $\xrightarrow[]{A}$ \lvbabbr{} $\xrightarrow[]{B}$ \itaabbr{}}\\
    & \small{BLEU / chrF++} & \small{BLEU / chrF++}\\
    \hline
    \circled{N4}/\circled{N4} & \lbs{70.57} / \lbs{82.56} & \lbs{64.19} / \lbs{81.26} \\
    \circled{N4}/\circled{R4} & 58.57 / 74.50 & 47.16 / 72.09 \\
    \circled{N4}/\circled{L4} & 63.90 / 78.09 & 59.47 / 78.46 \\\hline
    
    \circled{R4}/\circled{N4} & 70.80 / 82.20 & 68.38 / 83.00 \\
    \circled{R4}/\circled{R4} & \obs{80.12} / \obs{88.94} & \obs{68.51} / \obs{84.73} \\
    \circled{R4}/\circled{L4} & 70.36 / 81.98 & 67.41 / 82.68 \\\hline
    
    \circled{L4}/\circled{N4} & 63.72 / 77.53 & 57.02 / 76.54 \\
    \circled{L4}/\circled{R4} & 57.13 / 73.32 & 46.95 / 71.52 \\
    \circled{L4}/\circled{L4} & \lbs{72.31} / \lbs{83.69} & \lbs{65.74} / \lbs{82.02} \\ 
    \end{tabularx}
  \caption{\label{tab:results-rtt}  
  Results for Round-Trip Translations
  }
\end{table}

\section{Related Work}
\label{sec:related-work}

Data augmentation such as back-translation~\cite{Senn:etal:2016,Hoan:etal:2018} and transfer learning~\cite{Zoph:etal:2016} are established strategies to improve MT systems. These concepts are discussed in~\citet{Hadd:etal:2022,Rana:etal:2023}, with a focus on low-resource scenarios.
The fact that the synthesised data plays a critical role in the quality of the systems trained on it, as it also introduces a certain degree of noise, was discussed extensively in~\citet{Edun:etal:2018,Xu:etal:2022}. It was shown that tagging synthetic data can be beneficial in the training process~\cite{Casw:etal:2019}. In our work we do not apply advanced techniques to differentiate synthetic from real translations in training.
The fact that RBMT systems can still be valuable for low-resource languages and can even help to achieve better results was also demonstrated for Northern Sámi~\cite{Aula:etal:2021} .
In our experiments, we could also observe this in the translation direction \latoit{}.
This could be due to the ability of the RBMT system to provide general knowledge that is not available in the relatively limited parallel training datasets~\cite{Aula:etal:2021}.
The use of LLMs for MT and different prompting techniques was investigated in~\citet{Zhan:etal:2023} and
their performance in the machine translation of low-resource languages has already been analysed in~\citet{Mosl:etal:2023}.
Even if they struggle to generate texts in low-resource languages~\cite{Robi:etal:2023}, 
it has already been claimed that they can contribute to advances in machine translation of such languages.
Our work is an example of how LLMs can be used in machine translation of a low-resource language; however, further prompt engineering is needed to make better use of such models.

\section{Conclusion}
\label{sec:conclusion}

In this work, we conducted a detailed comparison of RBMT, NMT and LLMs for back-translation in a low-resource scenario. 
We have tested various back-translation approaches and evaluated them for a previously unexplored language in the field of machine translation.

Our current methodology involved the exclusion of numerous Ladin monolingual sentences.
However, this filtering would be less important for the translation direction \latoit{}. 
This previously discarded data could be re-incorporated to improve the performance of the models in this particular translation direction.

The round-trip translation scores indicate that the initial back-translation with the RBMT system leads to more robust models. Improving the ambiguity resolution of this rule-based translation system could lead to even better results.

The simplicity of the prompts used to feed the LLMs provides a further starting point for investigations.
In particular, the question arises as to whether the results can be improved by further prompt engineering, e.g., by including the meaning for the distinct words occurring in a text using the available dictionary. 
Investigating the effects of prompt optimisation could provide new insights into maximising the efficiency of LLMs in machine translation, especially in low-resource scenarios.

We plan to address these research questions in our future work.

\section*{Acknowledgements}

This research was conducted in collaboration with the Ladin Cultural Institute \emph{Micurà de Rü} and funded by the \emph{Regione Autonoma Trentino-Alto Adige/Südtirol}.

\bibliography{acl_latex}

\appendix

\newpage
\onecolumn
\section{Prompt template}
\label{sec:appendix-llm}
  \begin{lstlisting}[style=json, label=lst:llm-prompt, caption=Prompt template used to obtain the Italian translations from the LLM., captionpos=b]
    I'll give you samples for the translation from Ladin to Italian:

    {
        "translations": [
            {
              "Ladin": "scrí sües minunghes", 
              "Italian": "scrivere le proprie opinioni"
            },
            {
              "Ladin": "mëte la secunda", 
              "Italian": "mettere la seconda"
            },
            {
              "Ladin": '"zessa, i á prescia!"', 
              "Italian": '"scansati, ho fretta!"'
            },
            {
              "Ladin": "passé ia le rü", 
              "Italian": "oltrepassare il fiume"
            },
            ...
            {
              "Ladin": "chësc liber é to", 
              "Italian": "questo libro è tuo"
            },
        ]
    }
    
     Please generate the translation of each of the  16 entries in the given dictionary, where the translations are empty. Return the same JSON dictionary where the values for Italian are filled:
    
    {
        "translations": [
          {
            "Ladin": "Sperun da salvé almanco val', dijun:", 
            "Italian": ""
          },
          {
            "Ladin": "Ince tröc toponims y cognoms ladins desmostra che l'identité ladina é coliada ala natöra y ala cultura da munt",
            "Italian": ""
          },
          {
            "Ladin": "De profesciun este pech... co este pa rové pro chësc laur?",
            "Italian": ""
          },
          ...
          {
            "Ladin": "I dormi n pü' domisdé y spo ciamó val' ora dan mesanöt",
            "Italian": ""
          }
        ]
    }
\end{lstlisting}
\end{document}